\documentclass{article}

\usepackage[dblblindworkshop, final]{neurips_2026}
\workshoptitle{AI for Stochastic Dynamics}

\usepackage[utf8]{inputenc}
\usepackage[T1]{fontenc}
\usepackage{amsmath,amssymb}
\usepackage{booktabs}
\usepackage{multirow}
\usepackage{float}
\usepackage{graphicx}
\usepackage{microtype}

\usepackage[hidelinks]{hyperref}
\usepackage{url}

\hypersetup{pdftitle={},pdfauthor={},pdfsubject={},pdfkeywords={},pdfcreator={},pdfproducer={}}

\usepackage{fancyhdr}
\fancypagestyle{workshophead}{%
  \fancyhf{}%
  \fancyhead[C]{\footnotesize 40th Conference on Neural Information Processing Systems (NeurIPS 2026). Workshop: AI for Stochastic Dynamics.}%
  \fancyfoot[C]{\thepage}%
  }
\title{Retention You Can Predict, Scope and Delete:\\ a $(\mu,\gamma,T)$ Law and a Structural Deletion Guarantee\\ for a Physics-Structured Memory Store}

\newcommand{\aff}[1]{\textsuperscript{\normalfont #1}}

\author{%
  Pratik Jawahar\aff{1,2}\thanks{Corresponding author: \texttt{pratik.jawahar@cern.ch}}
  \And Riccardo Maggioni\aff{3}
  \And Jonas Petersen\aff{3}
  \And Matteo Cozzi\aff{3}
  \And Gian-Alessandro Lombardi\aff{2}
  \And Marek Gajewski\aff{2}
  \And Camilla Mazzoleni\aff{3}
  \And Federico Martelli\aff{2}
  \And Maurizio Pierini\aff{1}
  \AND
  \normalfont \aff{1}CERN \quad \aff{2}Forgis AG \quad \aff{3}ETH Zurich
}

\begin{document}
\maketitle
\thispagestyle{workshophead}
\begin{abstract}
Deployed memory systems enforce retention with explicit bookkeeping: decay factors, expiry timestamps, or learned forget gates decide when an entry expires.
However, these rules do not specify the rate at which a stored value degrades, how tuning a retention parameter changes that rate, or what state the store keeps after an item is removed.
In this work, we treat forgetting as a dynamical property of a memory store built as a Latent Dynamics Model, in which a damped symplectic step of a learned Hamiltonian advances the latent state. A computable budget of three variables then sets retention: the stored direction's spectral mass $\mu$, the damping $\gamma$, and the temperature $T$. We show three properties. First, the retention dial has a computable critical point: the half-life is non-monotonic in friction and forms a V-curve minimized near $\gamma_{\rm crit}\approx2\varepsilon\mu$, the small-step limit of the exact critical point; on learned units the measured minima sit at $\approx0.90\,\gamma_{\rm crit}$. The curve spans approximately 11 orders of magnitude in $\mu^2$ and holds for both designed and emergent (learned-MLP) units. Second, retention can be scoped: at $T>0$, friction preserves memory while temperature erases it, so a local friction hole acts as a memory vault, with a $107.77\pm4.78\times$ retention factor on designed units. Third, removal is structural: under canonical \citet{blelloch_strongly_2007} placement, store-level deletion is exact. Under stated operational conditions, the store after deletion is byte-identical to a store that never held the item, on a designed, non-learned 3-dimensional datastore at capacities 8--64.
\end{abstract}

\section{Introduction}\label{sec:intro}

Deployed memory stores implement retention and deletion with a decay factor, an invalidation timestamp, a learned forget gate, or an explicit \texttt{DELETE} call. Forgetting nevertheless remains a failure mode that standard benchmarks often do not measure adequately \citep{yang_control-plane_2026,uddin_recall_2026}. A programmatic policy is not a dynamical law: one cannot read the half-life of a stored value off a trained LSTM, identify the mechanism that tunes it, or verify the residue left once a value is removed.

The gap is widest at deletion. Agent-memory research supplies mechanisms for writing, consolidating, and retrieving data, but deletion relies on bookkeeping, such as a timestamp or a dropped row \citep{rasmussen_zep_2025,chhikara_mem0_2025}, which leaves measurable residue after the entry is nominally deleted \citep{chakraborttii_ghost_2026,wang_memleak_2026}. Sec.~\ref{app:related} discusses related work.

In this work, we make forgetting a prescribed dynamical property of the store, set by parameters that can be chosen, targeted, and analyzed in closed form, rather than a side effect or an external bookkeeping rule. Modeling the store as a physical system gives computable answers to three questions:
\begin{enumerate}
    \item What is the retention policy, and where is its critical point? (Sec.~\ref{sec:vcurve})
    \item Can retention be scoped to a single local item? (Sec.~\ref{sec:vault})
    \item Is structural deletion exact, and what information still leaks? (Sec.~\ref{sec:deletion})
\end{enumerate}

We first review related work (Sec.~\ref{app:related}) and define the store and its update step (Sec.~\ref{sec:setup}). We then answer the three questions in turn (Sec.~\ref{sec:results}) and close with limitations (Sec.~\ref{sec:limits}).

\section{Related Work}\label{app:related}

We first review retention policies in deployed memory systems. Agent memories are external stores that a model writes to, reads from, and edits (e.g., MemGPT, \citealp{packer_memgpt_2024}; Mem0, \citealp{chhikara_mem0_2025}). Other designs keep a bounded compressive state inside the network (e.g., Infini-attention, \citealp{munkhdalai_leave_2024}). Their retention policies rely on scores and gates. Examples include a $0.995$ recency decay factor in a ranking heuristic \citep{park_generative_2023}, Ebbinghaus decay of memory strength \citep{zhong_memorybank_2023}, learned per-memory expiration spans \citep{sukhbaatar_not_2021}, and forget gates $\alpha_t$ optimized at test time \citep{behrouz_titans_2024}. The learned gate in Titans \citep{behrouz_titans_2024} is structurally similar to the retention dial of Sec.~\ref{sec:vcurve}. Still, none of these works states a physical retention law, gives a half-life, defines a read tolerance, or prices the information loss.

Deletion mechanisms exist in deployed systems and in formal work. Deployed systems handle deletion by invalidation. Zep invalidates contradicted edges with timestamps but keeps the latent record \citep{rasmussen_zep_2025}, and Mem0 drops rows from its active context. Byte-identity matters for security, however: soft-deleted vectors in graph ANN databases remain reconstructible from raw index files \citep{chakraborttii_ghost_2026}. Recent benchmarks find that real-world deployment failures are mostly forgetting failures \citep{yang_control-plane_2026}, and that outdated retention drives significant recommendation errors \citep{uddin_recall_2026}. Formally, \emph{certified} removal is \citeauthor{guo_certified_2023}'s \citeyearpar{guo_certified_2023} Sec.~2, Eq.~(1), an $\varepsilon$ condition followed by an unnumbered $(\varepsilon,\delta)$ relaxation. Exact methods delete by isolation, and their cost and capacity have been formalized \citep{bourtoule_machine_2019,ginart_making_2019,sekhari_remember_2021}. Our work sits between these approaches: it gives a store-level, bit-exact structural guarantee from canonical geometric placement \citep{blelloch_strongly_2007}, not a stochastic system-level unlearning heuristic.

Prior work also studies forgetting laws in learned and physical systems. Learned recurrent models such as LSTMs \citep{hochreiter_long_1997} lack a stateable physical forgetting law. In biological and artificial networks, representational drift has been observed to reorganize unit representations over time \citep{aitken_geometry_2022}. We instead track one coset coordinate under defined physical dynamics. In theoretical physics, dissipation turns type-A Nambu--Goldstone modes from propagating into diffusive \citep{minami_spontaneous_2018}. Strictly equivariant flows also have Lyapunov neutral modes \citep{mo_symmetry-protected_2026}; these give kinematic protection, a gap complementary to our physical budget ($\mu^2$).

\section{The Store as a Damped Hamiltonian System}\label{sec:setup}

Methodologically, the store is a Latent Dynamics Model: a model that structures the dynamics of a latent space produced by an encoder. The latent state $(q,p)$ advances by a damped symplectic step of a learnable Hamiltonian $H=T(p)+V_\theta(q)$, which describes the dynamics of the encoder's latent space rather than of the input system itself. The damping $p\mapsto(1-\gamma)p$ supplies forgetting, and Langevin noise sets the temperature $T$. Items can optionally be written as \emph{atoms}: localized energy wells superposed into one energy function, which is read by dynamical relaxation. Our reference architecture is the Causal Learning Unit (CLU), introduced as CHLU by \citet{jawahar_chlu_2026}. Its core is exactly solvable: in the eigenbasis of the mass-whitened Hessian, one damped step acts on each normal mode as a $2\times2$ matrix fixed by $(\varepsilon\mu,\gamma)$, whose spectral radius sets that mode's retention.

We work under the following scope, assumptions, and constraints:
\begin{itemize}
    \item \emph{Scale.} These early results use a novel, designed store, not a deployed, large-scale LLM agent memory or a general system benchmark.
    \item \emph{Claim type.} We report a mechanism and the measured laws that bound it, not an end-to-end system-performance result.
    \item \emph{Nomenclature}, following CHLU~\citep{jawahar_chlu_2026}.
    \emph{Inertial mass ($M_i$):} the learned diagonal of the kinetic term; larger values give slower dynamics.
    \emph{Spectral mass ($\mu_k^2$):} $\lambda_k(M^{-1/2}\nabla^2V_\theta M^{-1/2})$; larger values give shorter retention. Retention claims always use $\mu$.
    \emph{Stored direction:} a written value occupies a near-flat coset direction.
    \emph{Integration step ($\varepsilon$):} fixed at $0.05$.
    \item \emph{Metrics.} Half-lives ($n_{1/2}$) are reported with their read tolerance ($\Delta$) and the thermal persistence length $\ell_\theta=\varepsilon\sqrt{T/M}/(\gamma r_*)$, the coset register's resolution limit at temperature $T$. A half-life predicts the decay envelope of the linearized mode; it is not a finite-time read-tolerance guarantee for the nonlinear learned dynamics or for arbitrary write amplitudes.
    \item \emph{Thermodynamic consistency.} $T>0$ requires the fluctuation--dissipation-consistent noise scale $\sigma^\star_i=\sqrt{M_iT\gamma(2-\gamma)}$ with a Newtonian kinetic mode.
\end{itemize}

\subsection{The Update Step}\label{app:derivation}
Every closed form in Sec.~\ref{sec:results} follows from a single object: the damped symplectic step, linearized about the vacuum. We define it here and give each derivation's key steps beside the result it supports; no result here is new.
\begin{itemize}
\item \emph{The step.} One update is: half-kick $p\mapsto p-\tfrac{\varepsilon}{2}\nabla V_\theta(q)$; drift $q\mapsto q'=q+\varepsilon M^{-1}p$; half-kick at $q'$; damping $p\mapsto(1-\gamma)p$, position-gated by the further factor $\bigl(1-\gamma_\phi(q')\bigr)$ evaluated at the \emph{updated} position when a friction field is present; and, at $T>0$, additive noise $p\mapsto p+\sigma\xi$, $\xi\sim\mathcal N(0,I)$, in the Newtonian kinetic mode that the scope list above requires.
\item \emph{Linearization.} Retention statements expand $V_\theta$ to second order about the vacuum and diagonalize in the mass-whitened Hessian $M^{-1/2}\nabla^2V_\theta\,M^{-1/2}$, whose eigenvalues are the spectral masses $\mu_k^2$. Write $h=\varepsilon\mu$.
\item \emph{The coset scale $F$.} $F^2$ denotes the mass-weighted squared displacement of the latent state per unit of the stored coordinate: for a stored angle on a ring of radius $r_*$ with tied channel mass $M$, $F^2=M r_*^2$. Diffusion coefficients are per unit dynamical time $t=n\varepsilon$.
\end{itemize}

\subsection{The per-mode step matrix}\label{app:deriv:1} In a whitened normal-mode coordinate $(x,p)$ with $V=\tfrac12\mu^2x^2$, composing the four sub-steps above gives the linear map $\bigl(\begin{smallmatrix}x\\p\end{smallmatrix}\bigr)\mapsto J\bigl(\begin{smallmatrix}x\\p\end{smallmatrix}\bigr)$,
\[
J=\begin{pmatrix}1-\tfrac{h^2}{2} & \varepsilon\\[2pt] -(1-\gamma)\,\varepsilon\mu^2\bigl(1-\tfrac{h^2}{4}\bigr) & (1-\gamma)\bigl(1-\tfrac{h^2}{2}\bigr)\end{pmatrix},
\qquad
\operatorname{tr}J=(2-\gamma)\Bigl(1-\frac{h^2}{2}\Bigr),\quad \det J=1-\gamma .
\]
This is the $2\times2$ matrix of Sec.~\ref{sec:setup} that is fixed by $(\varepsilon\mu,\gamma)$ alone; its spectral radius sets each mode's retention.

\subsection{Conformal symplecticity at constant damping; volume contraction under the gate}\label{app:deriv:2} The three Verlet sub-steps are shears (each Jacobian is triangular with unit diagonal), so they are symplectic and their composition has determinant $1$. The damping sub-step $q\mapsto q$, $p\mapsto(1-\gamma)p$ scales the symplectic form by $(1-\gamma)$, i.e.\ $J^\top\Omega J=(1-\gamma)\Omega$, and contributes $(1-\gamma)^d$ to the determinant. With constant damping the step is therefore conformally symplectic, with $\det J=(1-\gamma)^d$. The gate multiplies each momentum component by the scalar $\bigl(1-\gamma_\phi(q')\bigr)$. Because $q'$ is already fixed when the gate acts, its Jacobian is block-triangular and contributes $\bigl(1-\gamma_\phi(q')\bigr)^d$ to the determinant, the position-gated form quoted in Sec.~\ref{sec:vcurve}. For this position-dependent damping we prove only the determinant formula, i.e.\ volume contraction; we do not show that the gated step is conformally symplectic.

\section{Results}\label{sec:results}

\subsection{The Retention Dial Has a Computable Critical Point}\label{sec:vcurve}

Damping acts as the store's retention dial, and one might expect more friction to shorten memory monotonically. However, the half-life is non-monotonic in friction, with a minimum fixed by the stored direction's spectral mass. The turning point can therefore be predicted in closed form rather than tuned empirically.

We use the update step of Sec.~\ref{sec:setup} on checkpoints with dimension 4 and hidden size 64, trained for 150 epochs at $\varepsilon=0.05$. We measure both \emph{designed units} (an $SO(2)$-invariant $V_\theta$, 5 seeds) and \emph{emergent units} (an MLP $V_\theta$ trained on the same symmetric data, 3 seeds).

\begin{figure}[htbp]\centering
\includegraphics[width=\linewidth]{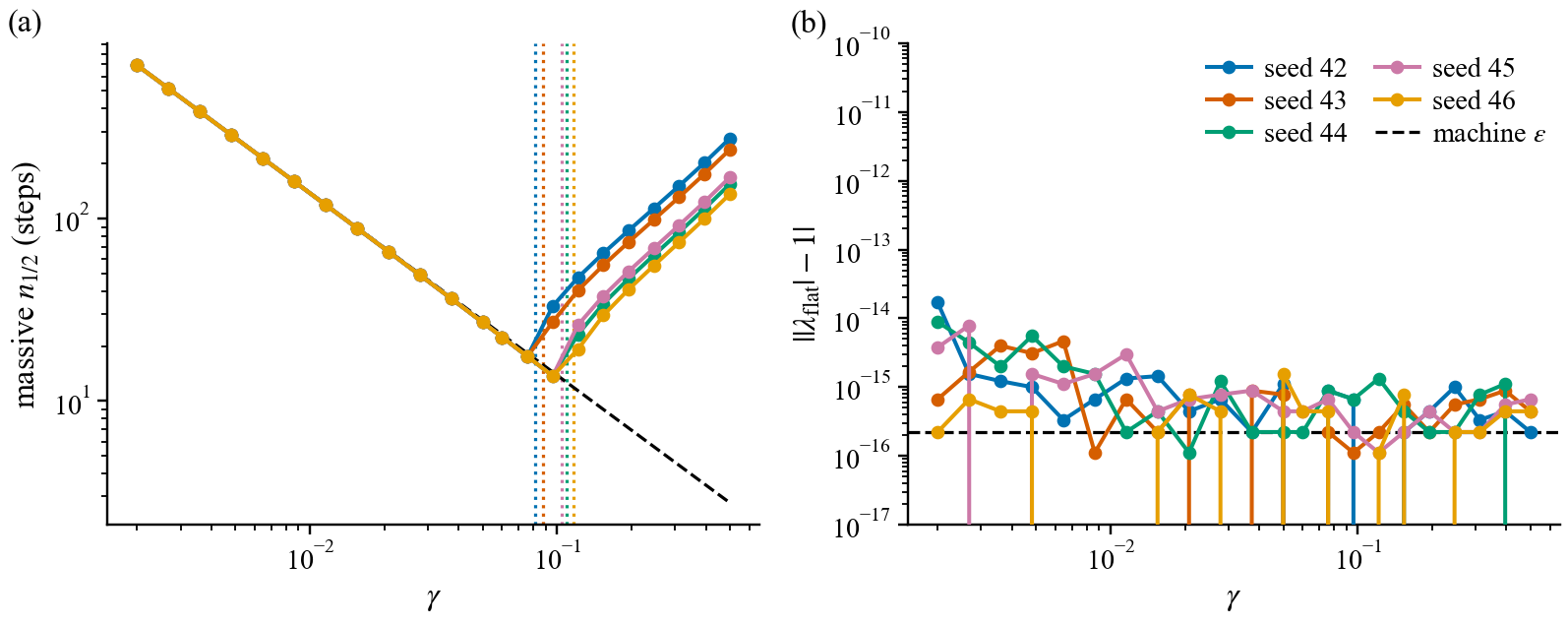}
\caption{The two regimes, read off the exact one-step Jacobian at the vacuum. (a) The massive radial mode is non-monotone in $\gamma$; each seed's minimum sits near its own $\gamma_{\rm crit}\approx2\varepsilon\mu$ (dotted), the small-step limit of the exact critical point, against the dashed underdamped envelope $2\ln2/\gamma$ on the same axes. (b) The flat mode holds $|\lambda|=1$ at every $\gamma$ to machine precision (dashed: machine $\varepsilon$): friction cannot erase the latch. Colors identify seeds in both panels. \emph{Five trained designed seeds, $T=0$, dim 4, hidden 64, $\varepsilon=0.05$, single-CPU: verification.}}
\label{fig:massiveflat}
\end{figure}

The latch, the overdamped register, and the underdamped working memory are three regimes of one curve, read at two values of $\mu$. On a trained designed $SO(2)$ vacuum, the massive radial mode's half-life $n_{1/2}(\gamma)$ forms a V-curve minimized near $\gamma_{\rm crit}\approx2\varepsilon\mu_{\rm rad}$, the small-step limit of the exact critical point (Fig.~\ref{fig:massiveflat}, Sec.~\ref{app:deriv:3}); the measured minima lie at $0.076$--$0.096$ against $0.082$--$0.116$ from this formula. The overdamped branch follows the $\mu^{-2}$ law until $\varepsilon\mu\approx\gamma/2$ and then saturates at a mass-independent floor. The log-slopes are $-1.006$ and $+1.23$ to $+1.27$ across the 5 seeds. The flat coset (Hessian $\mu^2\approx10^{-15}$) lies on the same curve in the limit $\mu\to0$, where $\gamma_{\rm crit}\to0$.

For the designed V-curve, the per-seed spectral masses are $\mu^2_{\rm rad}=0.670,0.771,1.190,1.093,1.348$, and the underdamped branch sets a mass-independent floor of $27.03$ steps at $\gamma=0.05$.

\phantomsection\label{app:deriv:3}We now derive the V-curve's two branches and $\gamma_{\rm crit}\approx2\varepsilon\mu$. The eigenvalues of Sec.~\ref{app:deriv:1} are $\lambda_\pm=\tfrac12\bigl(\operatorname{tr}J\pm\sqrt{\operatorname{tr}^2J-4\det J}\bigr)$, and the discriminant's sign splits the curve.

\emph{Underdamped branch} ($\operatorname{tr}^2J<4\det J$): the pair is complex with $|\lambda|^2=\det J=1-\gamma$, so
\[
n_{1/2}=\frac{\ln 2}{-\tfrac12\ln(1-\gamma)}\;\xrightarrow{\ \gamma\ll1\ }\;\frac{2\ln2}{\gamma},
\]
independent of $\mu^2$ and of the checkpoint --- the left-branch result of App.~\ref{app:emergent} ($\lambda_{\rm ret}=\sqrt{1-0.002}=0.998999499$, $n_{1/2}=692.5$ vs.\ $2\ln2/0.002=693.1$), the dashed $2\ln2/\gamma$ envelope of Fig.~\ref{fig:massiveflat}a, and, evaluated at $\gamma=0.05$, the mass-independent floor of $27.03$ steps of the designed V-curve above.

\emph{Overdamped branch}: substituting $\lambda=1-\delta$ into $\lambda^2-\lambda\operatorname{tr}J+\det J=0$ gives exactly $\delta^2-\bigl[\gamma+(2-\gamma)\tfrac{h^2}{2}\bigr]\delta+(2-\gamma)\tfrac{h^2}{2}=0$, whose retained (slow) root is $\delta=(2-\gamma)h^2/(2\gamma)\,\bigl(1+O(h^2/\gamma^2)\bigr)$. Hence
\[
n_{1/2}=\frac{\ln2}{-\ln(1-\delta)}\;\simeq\;\frac{2\ln2\,\gamma}{(2-\gamma)\,\varepsilon^2\mu^2},
\]
the $\mu^{-2}$ law of Sec.~\ref{sec:vcurve}. Its log-slope is $d\ln n_{1/2}/d\ln\gamma=1+\gamma/(2-\gamma)$: it approaches the $+1$ asymptote only as $\gamma\to0$ and exceeds it at every finite $\gamma$, while the underdamped slope $-\gamma/[(1-\gamma)(-\ln(1-\gamma))]=-1-\tfrac{\gamma}{2}+O(\gamma^2)$ stays close to $-1$. This is why the two measured branches are asymmetric (slopes near $-1$ below, $+1.1$ to $+1.3$ above, per the fit windows of this section and App.~\ref{app:emergent}).

\emph{The critical point}: the branches meet where $\operatorname{tr}^2J=4\det J$, i.e.\ $(2-\gamma)(1-h^2/2)=2\sqrt{1-\gamma}$, which $h^\ast(\gamma)=\bigl(1-\sqrt{1-\gamma}\bigr)\sqrt{2/(2-\gamma)}$ solves exactly. Inverting at small $h$ gives $\gamma^\ast=2\varepsilon\mu\bigl(1+O(\varepsilon\mu)\bigr)$. The value $\gamma_{\rm crit}=2\varepsilon\mu$ used in the main text is the small-step limit of this exact critical point, not the exact discrete value. On the emergent seeds the measured minima sit at $\approx0.90\,\gamma_{\rm crit}$ (App.~\ref{app:emergent}); the leading-order formula does not account for this gap. A flat coset is the $\mu\to0$ limit of the same curve, with $\gamma_{\rm crit}\to0$: the unification claim of Sec.~\ref{sec:vcurve}.

\begin{figure}[htbp]
  \centering
  \includegraphics[width=0.6\linewidth]{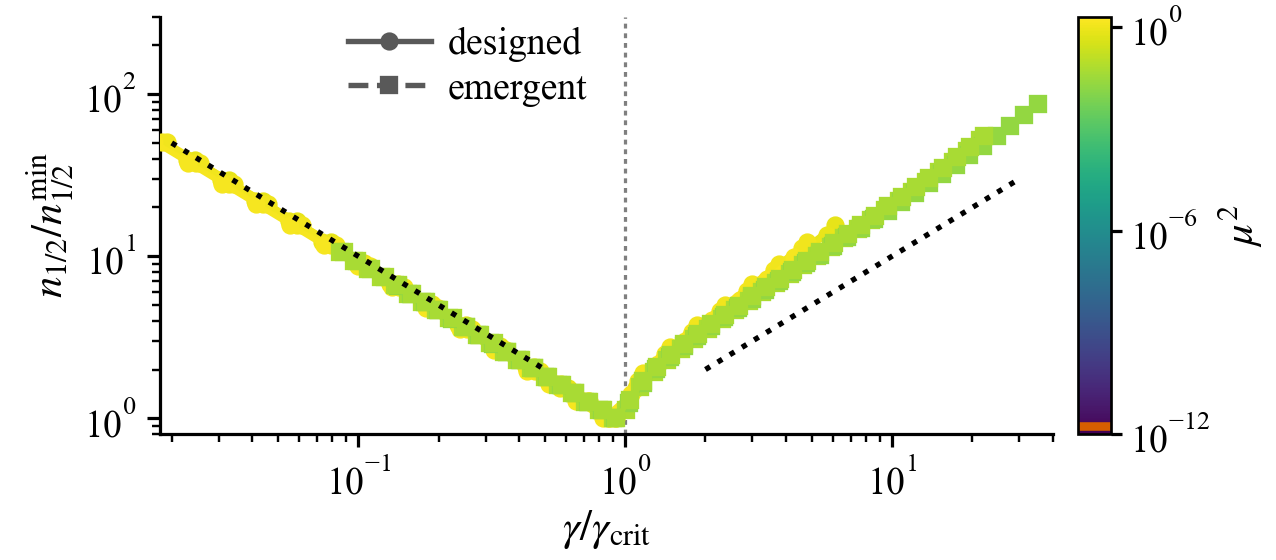}
  \caption{\emph{The retention dial and its critical point, collapsed.} Eight measured curves collapse onto one: five designed radial modes (circles, verification, $\mu^2_{\rm rad}=0.670$--$1.348$) and three emergent learned-MLP coset modes (squares, evidence, $\mu^2_{\rm soft}=2.0$--$5.4\times10^{-2}$, 3/3 seeds), against the dotted slope-$\mp1$ asymptotes. The vertical line marks $\gamma=\gamma_{\rm crit}$, and color gives $\mu^2$ (the colorbar tick is the probe floor). The designed flat coset is the $\mu\to0$ corner of the same curve.}
  \label{fig:collapse}
\end{figure}

The law transfers to emergent units. On three emergent MLP checkpoints, we read the $T=0$ half-life on the coset-tracked, near-flat stored direction that holds a written value, not on the stiff radial mode above. It traces the same V-curve on all 3 seeds (Figs.~\ref{fig:collapse} and \ref{fig:lambdacoset}). Its minimum sits at $0.902\pm0.003\,\gamma_{\rm crit}$, a measured gap that the leading-order formula does not account for; Table~\ref{tab:numbers} gives the asymptotic log-slopes. Across both families, the curve holds over eleven orders of magnitude in $\mu^2$ (Table~\ref{tab:numbers}, Fig.~\ref{fig:collapse}). The low end of that span is the ring-profile probe's resolution on a checkpoint whose Hessian $\mu^2$ is machine zero rather than a spectral mass, so the eleven orders are one curve on one instrument (probe-floor tick in Fig.~\ref{fig:collapse-full}). A designed unit and a learned unit follow the same curve, read at different spectral masses.

\begin{figure}[!htb]
\centering
\includegraphics[width=\linewidth]{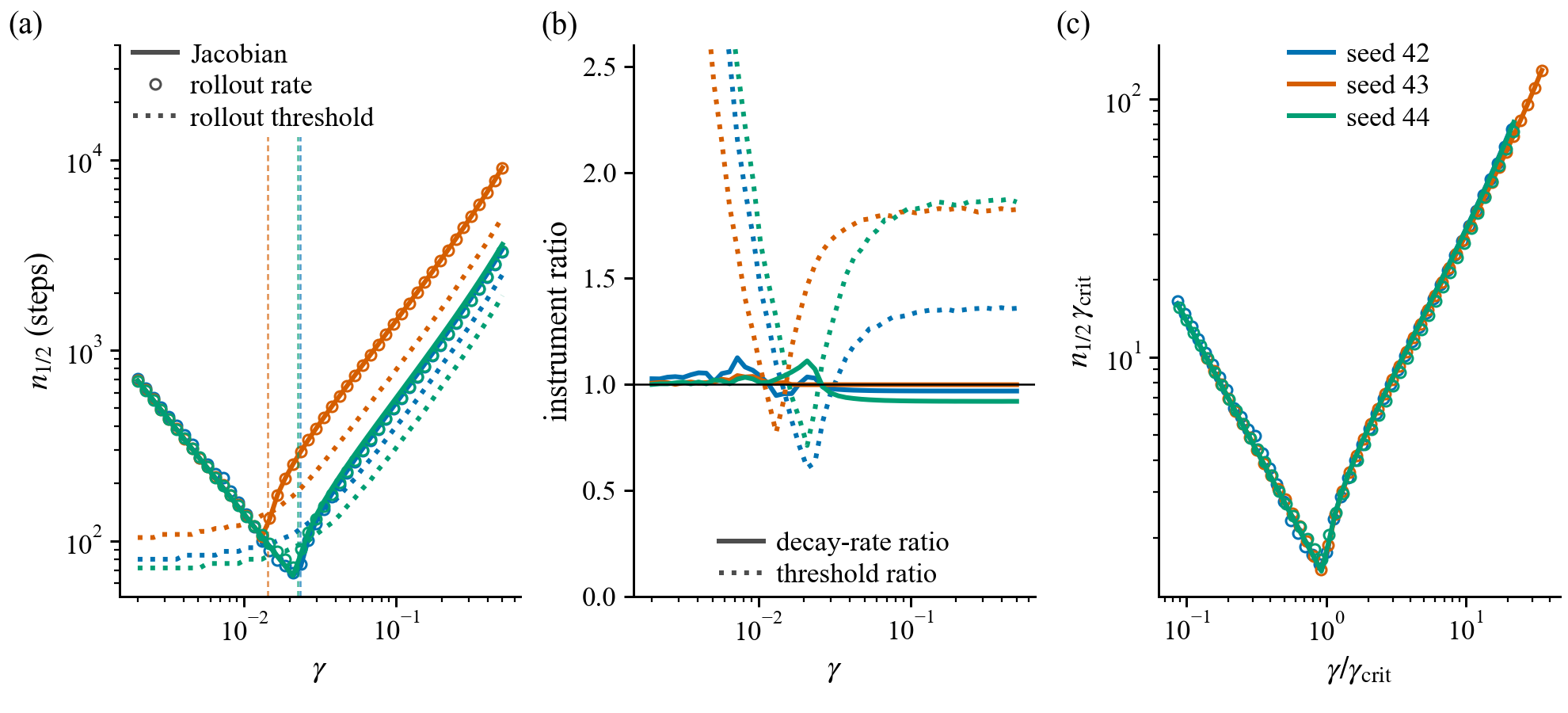}
\caption{The V-curve measured a second way. (a) $T=0$ $n_{1/2}(\gamma)$ per emergent seed on the one-step Jacobian (I-J, lines), the rollout envelope-rate instrument (I-R3, circles) and the rollout first-crossing threshold (I-R1, dotted), at $\delta=0.05$ rad; dashed vertical lines mark each seed's $\gamma_{\rm crit}$. (b) The per-$\gamma$ instrument ratio: the rate ratio (solid) stays near $1$ ($0.995\pm0.003$ over the asymptotic fit window) and falls slightly below it at the largest $\gamma$; the threshold ratio (dotted) is offset and flat in $\gamma$ above $\gamma_{\rm crit}$. (c) The same data in collapsed variables. Colors identify seeds in all panels. \emph{(Emergent MLP, 3 seeds, dim 4, hidden 64, $\varepsilon=0.05$, float64, $T=0$, pre-registered $\Rightarrow$ evidence.)}}
\label{fig:twoinstruments}
\end{figure}

The primary measurements use one-step Jacobians, so we re-measured the whole curve by nonlinear rollout, extracting $n_{1/2}$ from the decay envelope's rate. This second instrument reproduces the curve's shape (Fig.~\ref{fig:twoinstruments}), and its minimum, $0.9001\pm0.0052\,\gamma_{\rm crit}$, agrees with the Jacobian's to $0.35\%$ (Table~\ref{tab:numbers}, App.~\ref{app:emergent}).

\subsection{Scoping Retention: Friction Preserves, Temperature Erases}\label{sec:vault}

Scoping retention to a single entry requires changing the dial locally. At $T=0$, a designed coset latches indefinitely at every friction we tested (Fig.~\ref{fig:massiveflat}b; drift bound and $\gamma$ range in Table~\ref{tab:numbers}). At $T>0$, however, the coset decays by diffusion, and friction slows this decay.

\begin{figure}[htbp]
  \centering
  \includegraphics[width=\linewidth]{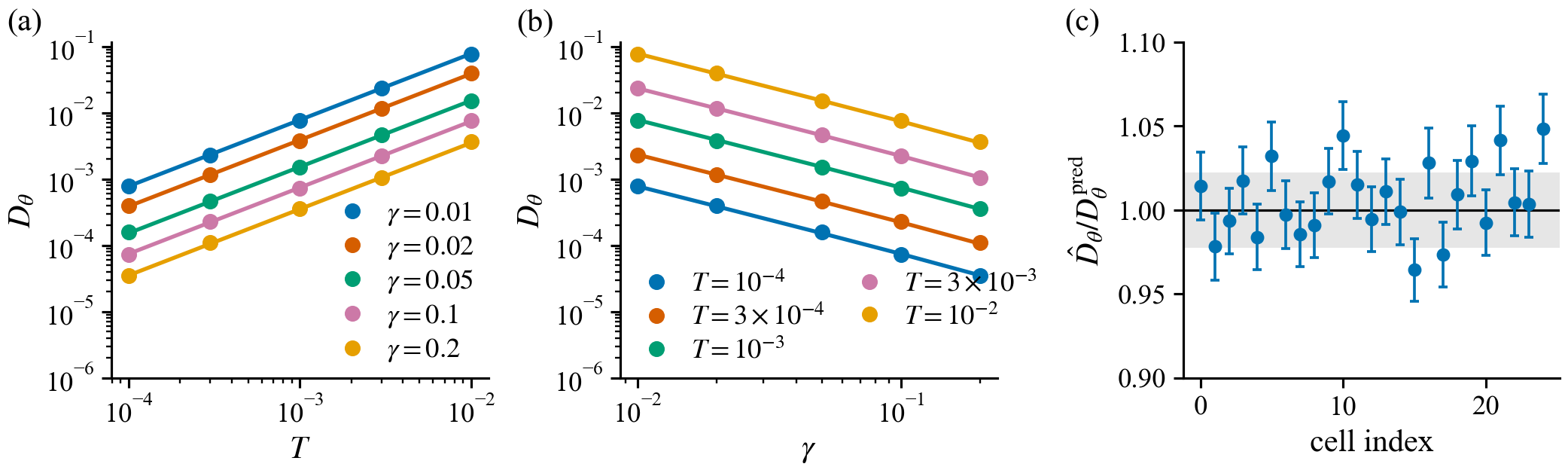}
  \caption{The coset diffusion law. In (a) and (b), markers are measured and lines are the law. (a) $D_\theta\propto T$ at five frictions; (b) $D_\theta\propto(2-\gamma)/\gamma$ at five temperatures; (c) the ratio $\hat D_\theta/D_\theta^{\rm pred}$ in each of the $25$ cells ($5\gamma\times5T$), against the law's value of $1$ (line); the shaded band shows the spread of the measured ratio about $1$. Over all cells the ratio is $1.0068\pm0.0219$. \emph{Multi-seed status: one trained designed-$SO(2)$ checkpoint (seed 44).} \texttt{langevin\_noise="fdt"}, Newtonian kinetic mode, $\Delta=0.5$ rad. \emph{Designed testbed: verification.}}
  \label{fig:dlaw}
\end{figure}

\begin{figure}[htbp]\centering
\includegraphics[width=\linewidth]{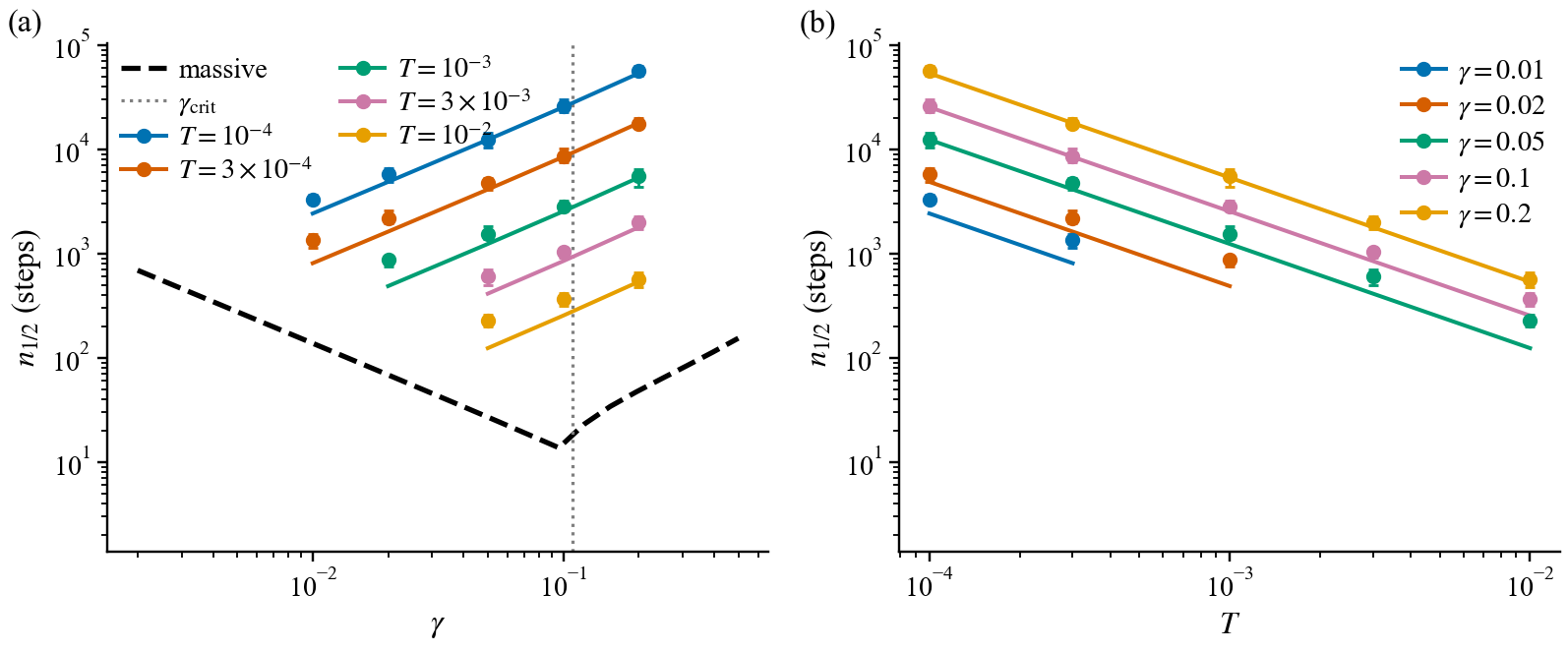}
\caption{The sign flip, and the lever that does erase. (a) The flat coset's half-life \emph{rises} with friction at every temperature, and the underdamped massive mode's (dashed) falls: the two stored directions respond to the dial with opposite sign ($\gamma_{\rm crit}=2\varepsilon\mu$ dotted). (b) Temperature is the forgetting lever, $n_{1/2}\propto1/T$ at every friction. Markers are measured half-lives with confidence intervals; solid lines are the diffusion-law prediction. \emph{Multi-seed status: one trained designed-$SO(2)$ checkpoint (seed 44); the $10/10$ seed $\times$ temperature statement and the $+0.9552\pm0.0422$ slope are this section's five-seed numbers, not this figure's.} \texttt{langevin\_noise="fdt"}, Newtonian kinetic mode, $\Delta=0.5$ rad. \emph{Designed testbed: verification.}}
\label{fig:signflip}
\end{figure}

We model the diffusion coefficient of the stored coordinate as $D_\theta=\varepsilon T(2-\gamma)/(2F^2\gamma)$, where the coset scale $F^2=M r_*^2$ is the mass-weighted squared displacement of the latent state per unit stored coordinate (Sec.~\ref{app:deriv:6}), and we verified it over a grid of $(\gamma,T)$ cells on the designed testbed (Table~\ref{tab:numbers}). Hence $n_{1/2}\propto\gamma^{+1}$. This sign flip ($\partial n_{1/2}/\partial\gamma>0$) holds in every seed and temperature combination we evaluated (Fig.~\ref{fig:signflip}; diffusion-law diagnostics in Fig.~\ref{fig:dlaw}). Only temperature erases memory ($\partial n_{1/2}/\partial T<0$).

\phantomsection\label{app:budget}The rest of this subsection follows one designed-testbed protocol. Its values come from the designed $SO(2)$ testbed, so they serve as verification. Every $T>0$ cell uses \texttt{langevin\_noise="fdt"} with a Newtonian kinetic mode; the legacy reference default violates these thermodynamic laws. Each reported half-life $n_{1/2}$ is quoted with its read tolerance $\Delta=0.5$ rad and its ratio $\ell_\theta/\Delta$.

\phantomsection\label{app:deriv:5}We first derive the discrete FDT scale $\sigma^\star$. The damping-plus-noise sub-step acts on each momentum component as the linear recursion $p'=(1-\gamma)p+\sigma\xi$, whose stationary variance solves $\mathrm{Var}=(1-\gamma)^2\mathrm{Var}+\sigma^2$, i.e.\ $\mathrm{Var}(p)=\sigma^2/[\gamma(2-\gamma)]$. Matching the Gibbs (equipartition) requirement $\mathrm{Var}(p_i)=M_iT$ of a Newtonian kinetic term forces
\[
\sigma^\star_i=\sqrt{M_i\,T\,\gamma\,(2-\gamma)}
\]
--- the thermodynamic-consistency scale of the scope list in Sec.~\ref{sec:setup}. For a relativistic kinetic term, the stationary momentum law of this linear recursion remains Gaussian while Gibbs demands a non-Gaussian marginal; this is why $T>0$ claims are restricted to the Newtonian mode.

\phantomsection\label{app:deriv:6}We next derive the coset diffusion law and the sign flip. Along a flat stored direction the force vanishes, so per step $\Delta\theta=\varepsilon\,p_\theta/(Mr_*)$, while $p_\theta$ follows the recursion of Sec.~\ref{app:deriv:5} with autocorrelation $\rho=1-\gamma$ and stationary variance $M T$. Summing $n$ correlated increments, $\mathrm{Var}\bigl(\sum_{k\le n}p_k\bigr)\to n\,MT\,\tfrac{1+\rho}{1-\rho}=n\,MT\,\tfrac{2-\gamma}{\gamma}$, hence $\mathrm{Var}(\theta_n)=\varepsilon^2Tn(2-\gamma)/(F^2\gamma)$, i.e.\ with $t=n\varepsilon$ and $\mathrm{Var}=2D_\theta t$:
\[
D_\theta=\frac{\varepsilon T(2-\gamma)}{2F^2\gamma}.
\]
A diffusive half-life to a read tolerance $\Delta$ scales as $n_{1/2}\propto\Delta^2/(\varepsilon D_\theta)\propto\gamma/[(2-\gamma)T]$. Increasing $\gamma$ therefore \emph{lengthens} coset memory ($n_{1/2}\propto\gamma^{+1}$, the sign flip of Sec.~\ref{sec:vault}), and only $T$ erases it ($n_{1/2}\propto T^{-1}$), while the massive mode of Sec.~\ref{app:deriv:3} responds with the opposite sign. These are the two rows of the $T>0$ budget table.

We test the diffusion law on 25 cells, a $5\gamma\times5T$ grid on seed 44. The ratio $\hat D_\theta/D_\theta^{\rm pred}$ is $1.0068\pm0.0219$, with a minimum of $0.9644$ and a maximum of $1.0484$. The slope $d\log D/d\log T$ takes the values $+0.998,+0.997,+0.992,+1.010,+1.006$. After isolating the discrete $(2-\gamma)$ factor, the slope $d\log D/d\log\gamma$ is $-1.014,-0.980,-0.999,-1.001,-0.982$.

We also measure the sign flip directly. At $T=10^{-3}$, we find $d\log n_{1/2}/d\log\gamma=+0.9552\pm0.0422$ across all 5 seeds, and the slope is positive in 10/10 conditions. Moving $\gamma:0.05\to0.2$ raises the half-life $3.77\pm0.23\times$ ($\ell_\theta/\Delta<0.05$; per-seed values $3.55,3.64,3.61,3.89,4.15$). Temperature acts in the opposite direction: $d\log n_{1/2}/d\log T=-0.956$ to $-0.979$ in the deep-diffusive regime.

\begin{table}[!htb]\centering\small
\caption{The $T>0$ face of the budget; all four entries are measured on the designed testbed.}
\begin{tabular}{p{0.19\linewidth}p{0.40\linewidth}p{0.31\linewidth}}
\toprule
Stored Direction & $\partial n_{1/2}/\partial\gamma$ (Increased Friction) & $\partial n_{1/2}/\partial T$ (Increased Heat)\\
\midrule
Coset / flat (designed) & $>0$, friction \emph{preserves} ($+0.955\pm0.042$; $3.77\times$ at $\gamma:0.05\to0.2$) & $<0$, temperature \emph{erases} ($-0.956$ to $-0.979$)\\
\addlinespace
Massive (underdamped) & $<0$, friction shortens (slope $-1.006$) & $\approx0$; $T$ sets only the fluctuation floor (max/min $\le1.013$)\\
\bottomrule
\end{tabular}
\end{table}

\phantomsection\label{app:deriv:4}We now derive the $T=0$ latch and show why $\gamma=0$ is not a latch. At $\mu=0$ the matrix of Sec.~\ref{app:deriv:1} is triangular, $J=\bigl(\begin{smallmatrix}1&\varepsilon\\0&1-\gamma\end{smallmatrix}\bigr)$, with eigenvalues $1$ and $1-\gamma$: for every $\gamma>0$ the stored coordinate carries $|\lambda_{\rm flat}|=1$ \emph{exactly}. Friction cannot erase the latch (Fig.~\ref{fig:massiveflat}b), and any residual momentum dies as $(1-\gamma)^n$. At $\gamma=0$, however, $J$ is a Jordan block for $\lambda=1$: residual momentum is never removed and the coordinate drifts ballistically, $\theta_n=\theta_0+n\,\varepsilon p_\theta/F^2$. This is the mechanism behind the $142.7$-rad $\gamma=0$ control reported below. With the confinement term $\alpha\|q\|^2$ present, every Hessian eigenvalue is lifted by $2\alpha$, so a flat direction becomes $\mu^2=2\alpha$ and Sec.~\ref{app:deriv:3}'s overdamped rate caps the lifetime at $\tau_{\max}=\Gamma/2\alpha$ (damping rate $\Gamma=\gamma/\varepsilon$).

On the designed units, we measure the $T=0$ latch directly. At absolute zero, $\big||\lambda_{\rm flat}|-1\big|\le1.7\times10^{-14}$ across 22 $\gamma$ values on 5 independent seeds. The drift is therefore $\le4.9\times10^{-12}$ rad per 200k integration steps in 30/30 cells. The $\gamma=0$ control, by contrast, drifts $142.7$ rad in only 20k steps.

\begin{figure}[htbp]\centering
\includegraphics[width=\linewidth]{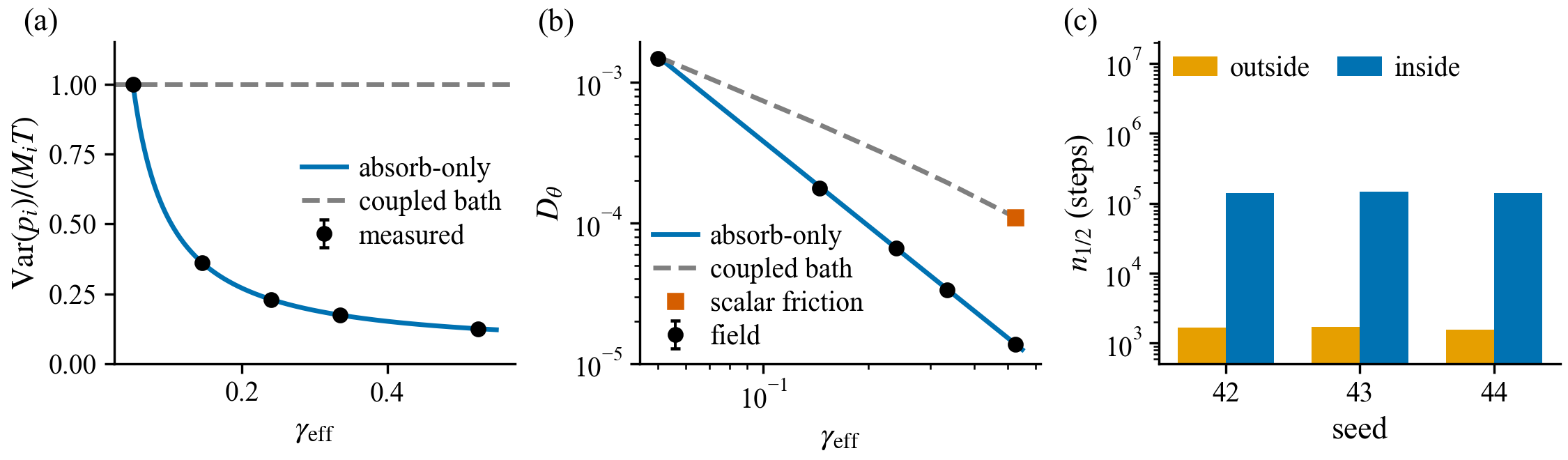}
\caption{The friction hole as a memory vault, on the designed arm (3 seeds; error bars are standard deviations over seeds). (a) The hole is a refrigerator: the momentum variance follows the absorb-only curve (solid) and rejects the coupled-bath prediction (dashed), giving $T_{\rm local}=1.26\times10^{-4}$ inside against $10^{-3}$ outside. (b) At the same $\gamma_{\rm eff}$, the diffusion constant differs by $8\times$ between a friction field (circles) and a scalar friction (square): the mechanism in one panel. Lines as in (a). (c) Per-seed first-passage half-lives outside and inside the hole at $T=10^{-3}$. These raw readings are boundary-layer biased on the outside arm, which is why their pooled ratio of $86.97\pm2.94\times$ is not quoted as the vault. The quoted vault factor, $107.77\pm4.78\times$, comes from the $\hat D_\theta$ estimator.}
\label{fig:vault}
\end{figure}

\emph{The memory vault.} We propose a memory vault: an absorb-only, position-gated friction field. A local hole in this field acts both as a brake, which increases dissipation, and as a refrigerator, which lowers the local effective temperature. On the designed arm, this gives a retention vault factor of $(\gamma_{\rm eff}/\gamma)^2$ (Fig.~\ref{fig:vault}), read from the $\hat D_\theta$ estimator and quoted against a uniform-scalar-friction control at matched $\gamma_{\rm eff}$. The direct first-passage reading is boundary-layer biased on the outside arm ($\ell_\theta/\Delta=0.079$), so it is not the quoted vault. Table~\ref{tab:numbers} lists all four readings (the $\hat D_\theta$ vault, the scalar control, the first-passage value, and the local temperature). The field also confines the register (Fig.~\ref{fig:vault-emergent}): the fraction of states outside the boundary ($|\theta|>1$ rad) drops to $0$ inside the hole (App.~\ref{app:vault}). The vault factor is how much longer a written value is retained inside the hole than outside it.

\phantomsection\label{app:deriv:7}We now derive the three roles of the friction hole: brake, refrigerator, and vault. Inside a uniform hole the step damps twice, $p'=(1-\gamma_\phi)(1-\gamma)p+\sigma^\star\xi$, so the effective damping is $\gamma_{\rm eff}=1-(1-\gamma)(1-\gamma_\phi)$, while the noise keeps the \emph{scalar}-$\gamma$ scale $\sigma^{\star2}=MT\gamma(2-\gamma)$ (absorb-only, no local re-emission). Repeating Sec.~\ref{app:deriv:5} with mismatched damping and noise gives
\[
\mathrm{Var}(p)=\frac{MT\,\gamma(2-\gamma)}{\gamma_{\rm eff}(2-\gamma_{\rm eff})}\equiv M\,T_{\rm local},
\qquad
T_{\rm local}=T\,\frac{\gamma(2-\gamma)}{\gamma_{\rm eff}(2-\gamma_{\rm eff})},
\]
which at $\gamma=0.05$, $\gamma_\phi\in\{0.1,0.2,0.3,0.5\}$ evaluates to the absorb-prediction column $0.36249/0.23082/0.17480/0.12591$ of the refrigerator table, and at $T=10^{-3}$ to $T_{\rm local}=1.26\times10^{-4}$. Repeating Sec.~\ref{app:deriv:6} with $\rho=1-\gamma_{\rm eff}$ gives
\[
D_\theta=\frac{\varepsilon T\,\gamma(2-\gamma)}{2F^2\gamma_{\rm eff}^2},
\qquad
\text{vault}=\frac{D_\theta(\text{no hole})}{D_\theta(\text{hole})}=\Bigl(\frac{\gamma_{\rm eff}}{\gamma}\Bigr)^{2}
\]
(every $(2-\gamma)$ factor cancels), $=110.25$ at $\gamma=0.05$, $\gamma_\phi=0.5$. The coupled-bath alternative instead rescales the noise to $\sigma^2=MT\gamma_{\rm eff}(2-\gamma_{\rm eff})$, giving $T_{\rm local}=T$ and $D_\theta=\varepsilon T(2-\gamma_{\rm eff})/(2F^2\gamma_{\rm eff})\propto\gamma_{\rm eff}^{-1}$, a vault of $(2-\gamma)\gamma_{\rm eff}/[\gamma(2-\gamma_{\rm eff})]=13.88$ at the same settings. The ratio of the two hypotheses is $\gamma_{\rm eff}(2-\gamma_{\rm eff})/[\gamma(2-\gamma)]=T/T_{\rm local}$: the $7.942\times$ separation factor of App.~\ref{app:vault} \emph{is} the refrigerator factor, identically.

On the designed arm, we measure $\mathrm{Var}(p_i)/(M_iT)=0.9986$ at baseline ($\gamma_\phi=0$), $0.36171\pm0.00129$ at ($0.1$), $0.23044$ at ($0.2$), $0.17513$ at ($0.3$), and $0.12600\pm0.00031$ at ($0.5$). The last value is close to the absorb-only prediction of $0.12591$. $T_{\rm local}$ thus drops to $1.26\times10^{-4}$, a $7.94\times$ refrigerator effect. The $D_\theta$-based vault is $8.42\times$, $22.39\times$, $44.31\times$ and $107.77\pm4.78\times$ at $\gamma_\phi=0.1/0.2/0.3/0.5$, matching the absorb-only prediction of $110.25$. The scalar control at the same $\gamma_{\rm eff}$ reads only $13.28\pm0.12\times$. At $T=0$ the hole erases nothing: drift stays $\le1.75\times10^{-12}$ rad per 200k steps, inside and outside the hole.

\subsection{Structural Deletion: Store-Level Guarantees, Lifetime Trade-offs, and Residual Leakage}\label{sec:deletion}

Standard architectures delete an item with best-effort heuristics, such as dropping an active row or writing a tombstone. When items are superposed into one energy function, however, standard placement becomes path-dependent. We therefore test whether the store's state can be a function of its live set alone, i.e., whether the layout depends only on which items are currently stored, not on the order or history of the writes that stored them.

To this end, we place items canonically, so that placement is a function of the live records and the fixed geometry alone. With \citeauthor{blelloch_strongly_2007}'s \citeyearpar{blelloch_strongly_2007} stable-matching table rules, store-level deletion is exact: with the waitlist engaged, it stays exact at every load we measured ($0.29\times$ to $1.71\times$ the lattice capacity; details below). Under explicit conditions (a budget sufficient for the cell count, zero baseline leakage, and priority- or attribute-based eviction; recency-based eviction is history-dependent and excluded), the store's layout is byte-identical to that of a store that never received the deleted item. The guarantee covers the isolated store, not the removal of information from a learned encoder or model, and our experiments do not measure end-to-end memory-system performance. The frozen encoder and any residue of past writes in a learned landscape are separate channels, and we claim no certified $(\varepsilon,\delta)$ unlearning.

We first place our deletion scheme in prior work. Uniquely represented (canonical) data structures date to \citet{snyder_uniquely_1977} and \citet{andersson_new_1995}; \emph{obliviousness}, a distributional condition on the memory representation that does not require canonical representation, is \citeauthor{micciancio_oblivious_1997}'s \citeyearpar{micciancio_oblivious_1997}; the weak and strong history-independence notions used here are \citeauthor{naor_anti-persistence_2001}'s \citeyearpar{naor_anti-persistence_2001}; the open-addressed realization is \citeauthor{blelloch_strongly_2007}'s \citeyearpar{blelloch_strongly_2007}, whose table is a stable matching between keys and slots under a global key priority, and we use that priority-greedy rule and that delete-time fix-up cascade; \citet{hutchison_uniquely_2008} had already taken unique representation into computational geometry. We claim no priority over order-independent placement and no novelty for the displacement rule or its delete-time repair. What is new is the composition: slots that are a lattice packing, so the canonical representation carries a minimum-separation certificate; content that is a continuously decaying amplitude whose decay commutes with deletion; a canonical object that is an energy function a dynamical read relaxes into; and a price that is negative where strong history independence is known to cost as much as an exponential slowdown relative to the weak notion for comparison-based structures \citep{goos_lower_2003}. ``Fix-up cascade'' is our name for the \texttt{NEXT}-chain delete-time repair of their Figure 1; the algorithm is theirs either way.

\phantomsection\label{app:deriv:8}We now show that decay commutes with deletion. Let the store state be the live key set $K$ with per-item amplitudes $\{a_i\}$, and let the layout be the canonical placement $L(K)$. Under the stated conditions (sufficient budget, priority/attribute eviction), this placement is a function of $K$ and the fixed geometry alone, and in particular independent of the amplitudes. Amplitude decay acts item-wise, $a_i\mapsto f(a_i)$, touching neither $K$ nor $L(K)$; deletion removes one key and recanonicalizes, touching neither the surviving amplitudes nor their decay clocks. The two operations act on disjoint components of the state $(K,\{a_i\})$, so they commute, and the post-delete store is byte-identical to the store built from the never-inserted history with the same elapsed decay. This is the commutation and byte-equality invoked in Sec.~\ref{sec:deletion}. Recency eviction makes $L$ a function of the insertion \emph{history} rather than of $K$, which is why it is excluded.

We test structural exactness at zero tolerance on all 24 insertion orders at $n=4$ and 200 random orders at $n\in\{8,16,40,64\}$. We also run 100 write/delete interleavings and deletes made mid-decay. In every case the minimum slot spacing is $1.540000$. On the shipped store, the centers, payloads, amplitudes and active mask are byte-equal, and the energy $V(q)$ is byte-equal across 64 random queries.

The packing price of canonical placement is negative. At the ceiling $K=64$, the canonical stable-matching rule admits $61/64$ offered sequential keys deterministically ($\sigma=0$; per-offered fraction $0.9531$). This rises to full capacity ($64/64$) when the lattice is inflated by $\times1.05$. The stochastic refuse-and-relocate allocator admits only $43/64$ on the same seeds (per-offered fraction $0.6719$). At the overflow limit without the waitlist, exactness fails, as expected: on 7 baseline cells receiving 8 capacity offers, $\mathrm{AUC}(n_{\rm live})$ is $1.00000\pm0.0000$ and the byte-equal fraction drops to $0.0000$.

\begin{table}[!htb]\centering\small
\caption{The laundering control, on the designed store with the shipped two-phase read geometry: 8 targets, 3 seeds, 128 paired environments per target, 18 amplitude levels.}
\begin{tabular}{p{0.34\linewidth}p{0.15\linewidth}p{0.19\linewidth}p{0.16\linewidth}}
\toprule
Evaluated Adversary Model & Physical Decay & TTL Flag Limit (Same Store) & Resolution Separation\\
\midrule
White-box architecture, fully exact & $1.000$ & $1.000$ & $0.000$\\
Query-level membership depth, fully exact & $0.983$ & $1.000$ & $0.017$\\
Query-level membership depth, $\sigma_{\rm obs}=0.1$ & $0.559$ & $0.996$ & $0.437$\\
\bottomrule
\end{tabular}
\end{table}

The store also shows a lifetime trade-off. In this implementation we observe a design trade-off among three properties: exact value fidelity, amplitude-independent address hold, and amplitude-independent read latency. The implemented store gives up amplitude-independent address hold, and two attempted repairs are refuted; we do not prove that the three are incompatible in general. The measurements behind this statement follow.
The implemented store gives up amplitude-independent address hold: its lifetime correlates with $a_i^2$ at $r=-0.85$. We refute both previously proposed repairs: multiplying payloads by their amplitudes breaks the value criterion at $A=1-\text{tol}/|a_i|$ (hard-death amplitudes $0.90/0.80/0.70/0.30$). \emph{Retrieval geometry:} $R_{50}$ contracts $1.146/1.083/0.979/0.874/0.752$ as $A:1\to0.5\to0.2\to0.1\to0.06$, a $1.52\times$ contraction from $1.146$ to $0.752$ that matches the pre-registered saddle prediction at all five points ($\pm0.20$); a TTL vector store's lookup radius is instead a constant hard step at $0.75$--$0.77$, independent of age. Below the barrier of its internal flag, the gated channel therefore contracts $R_{50}$ from $1.135$ to $0.771$. We make no deletion-cost statement, because canonical synchronization requires a full $O(n)$ rebuild per operation.

\section{Limitations and Future Work}\label{sec:limits}

We note the following limitations of our evidence and directions for future work:
\begin{itemize}
    \item \emph{Scale and architecture scope.} This work is limited to memory stores built as Latent Dynamics Models (e.g.\ CLU). Early experiments use dimension 4, hidden size 64 and a single CPU, across 5 designed and 3 emergent seeds. We infer nothing about larger scales and leave scaling to future work.
    \item \emph{Designed-symmetry precondition.} The exact $T=0$ retention latch currently requires continuous coset geometries that must be designed into the architecture.
    \item \emph{Thermodynamic scope.} The raw $T/\gamma$-exponents cannot distinguish designed from emergent units. Any $T>0$ evaluation requires \texttt{fdt} noise together with the Newtonian kinetic mode.
    \item \emph{Deletion scope.} The exact deletion guarantee holds only at the store level, under set-function eviction. Recency-based eviction breaks latent history independence and is therefore excluded.
    \item \emph{No task-level results.} This work presents a theoretical concept with empirical validation. We leave task-level benchmarks to future work.
    \item \emph{Future work.} We intend to test wider emergent collapses in $\mu^2$, remeasure the vault contrast, measure the amortized per-delete cost against flat datastores, and build localized temperature fields. Making these practical and performant at the required scales will let end-to-end Latent Dynamics Models acting on real data reorganize a learned latent space in a useful and controllable way.
\end{itemize}

\nocite{jude_capturing_2023}
\bibliographystyle{plainnat}
\bibliography{refs}

\appendix

\section{Values Quoted in the Main Text}\label{app:numbers}

Sec.~\ref{sec:results} states its results in words; this appendix collects the measured values behind them (Tab.~\ref{tab:numbers}), each beside the arm it was measured on and the rider it must be quoted with. The \emph{Arm} column gives the class of each measurement: verification means a designed testbed (an architecturally designed $SO(2)$ potential), and evidence means a learned system (an MLP $V_\theta$ trained on symmetric data). Each value repeats one from the section or appendix named beside it; none is new.

\begin{table}[!htb]\centering\footnotesize
\setlength{\tabcolsep}{4pt}
\caption{The measured values behind Sec.~\ref{sec:results}, with the arm each was measured on and the rider each must travel with. \emph{Arm}: \emph{verification} = designed testbed; \emph{evidence} = learned system. All rows come from the checkpoints and grids of Secs.~\ref{sec:vcurve}--\ref{sec:vault} and Apps.~\ref{app:emergent}--\ref{app:vault} (dim 4, hidden 64, $\varepsilon=0.05$, single-CPU); every $T>0$ cell uses \texttt{langevin\_noise="fdt"} with a Newtonian kinetic mode.}
\label{tab:numbers}
\begin{tabular}{p{0.165\linewidth}p{0.15\linewidth}p{0.15\linewidth}p{0.075\linewidth}p{0.28\linewidth}}
\toprule
Result & Quantity & Value & Arm & Scope\\
\midrule
\multirow{3}{\linewidth}{V-curve, emergent arm (Sec.~\ref{sec:vcurve}, App.~\ref{app:emergent})}
 & argmin, one-step Jacobian & $0.902\pm0.003\times\gamma_{\rm crit}$ & evidence & 3 emergent MLP seeds, $T=0$; a learned potential, not a designed one\\
 & log-slope, overdamped branch & $-1.0020\pm0.0003$ & evidence & same 3 emergent seeds\\
 & log-slope, underdamped branch & $+1.116\pm0.011$ & evidence & same 3 emergent seeds\\
\addlinespace
Collapse span (Sec.~\ref{sec:vcurve}) & $\mu^2$ range spanned by the one curve & $[1.7\times10^{-12},\,7\times10^{-2}]$ & both arms & Probe resolution: the low endpoint is the ring-profile probe's resolution on a checkpoint whose Hessian $\mu^2$ is machine zero (\emph{not} a measured spectral mass)\\
\addlinespace
\multirow{2}{\linewidth}{Cross-instrument check (App.~\ref{app:emergent})}
 & argmin, rollout envelope rate & $0.9001\pm0.0052\times\gamma_{\rm crit}$ & evidence & 3 emergent seeds, $\delta=0.05$ rad\\
 & argmin, one-step Jacobian & $0.9032\pm0.0027\times\gamma_{\rm crit}$ & evidence & same cells; the instrument gap is a level, not a rate\\
\addlinespace
$T=0$ latch (Sec.~\ref{sec:vault}, Sec.~\ref{app:budget}) & coset drift & $\le4.9\times10^{-12}$ rad / 200k steps & verifi\-cation & designed $SO(2)$ coset, every $\gamma\in[0.002,0.5]$; $T=0$ only\\
\addlinespace
Coset diffusion law (Sec.~\ref{sec:vault}, Sec.~\ref{app:budget}) & $\hat D_\theta/D_\theta^{\rm pred}$ & $1.0068\pm0.0219$ & verifi\-cation & 25 $(\gamma,T)$ cells on one trained designed checkpoint (seed 44), $\Delta=0.5$ rad\\
\addlinespace
\multirow{4}{\linewidth}{Friction-hole vault (Sec.~\ref{sec:vault}, App.~\ref{app:vault})}
 & $T_{\rm local}$ inside vs.\ outside & $1.26\times10^{-4}$ vs.\ $10^{-3}$ & verifi\-cation & designed arm, $\gamma=0.05$, $\gamma_\phi=0.5$; absorb-only field\\
 & vault factor, $\hat D_\theta$ estimator & $107.77\pm4.78\times$ & verifi\-cation & the quoted vault: designed $SO(2)$, 3 seeds, $T=10^{-3}$, $\Delta=0.5$ rad\\
 & scalar-friction control & $13.28\pm0.12\times$ & verifi\-cation & uniform scalar friction at matched $\gamma_{\rm eff}$, same cells\\
 & raw first-passage ratio & $86.97\pm2.94\times$ & verifi\-cation & boundary-layer biased on the outside arm ($\ell_\theta/\Delta=0.079$); \emph{not} the quoted vault\\
\bottomrule
\end{tabular}
\end{table}

\section{The Emergent Arm, and the V-Curve on a Second Instrument}\label{app:emergent}

\begin{figure}[!htb]
\centering
\includegraphics[width=\linewidth]{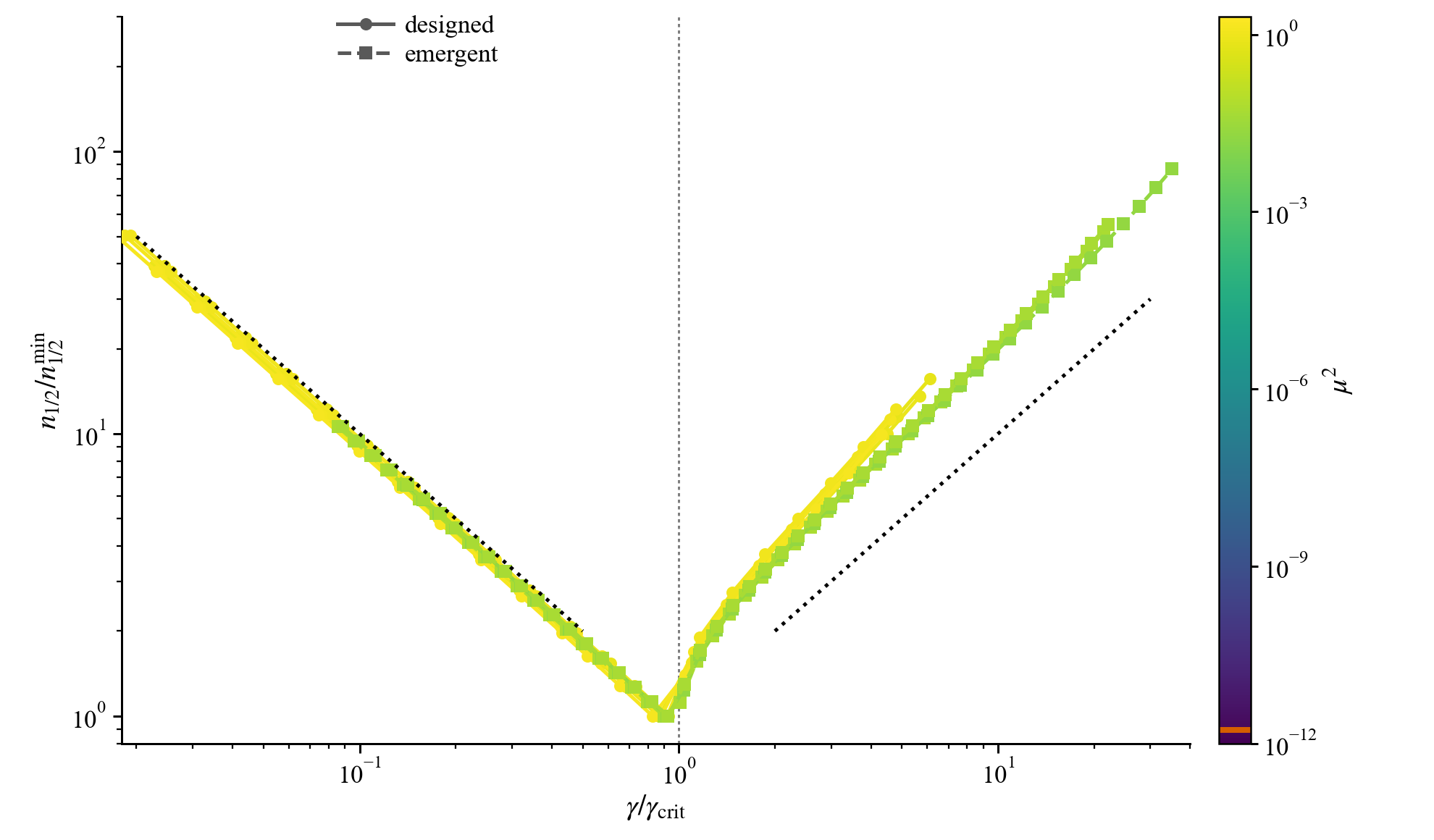}
\caption{\emph{The damping optimum, collapsed: the full-size version of Figure~\ref{fig:collapse}.} The same eight measured curves, five designed seeds (circles, verification) and three emergent seeds (squares, evidence), with the same dotted slope-$-1$ (underdamped) and slope-$+1$ (overdamped, $\propto\mu^{-2}$) asymptotes and the vertical line at $\gamma=\gamma_{\rm crit}=2\varepsilon\mu$. Color gives $\mu^2$ of the plotted mode; the vermillion tick on the colorbar marks the flat-coset probe floor ($1.7\times10^{-12}$). No plotted value differs from Figure~\ref{fig:collapse}. \emph{(One-step Jacobian; dim 4, hidden 64, $\varepsilon=0.05$, single-CPU; \texttt{langevin\_noise="fdt"}, Newtonian kinetic mode.)}}
\label{fig:collapse-full}
\end{figure}

We test the emergent MLP checkpoints on 3 seeds, alongside a matched designed control. All predictions below were pre-registered before any test harness was run, so the results count as evidence.

We use four instruments over the same fit windows ($\gamma<\gamma_{\rm crit}/2.5$ and $\gamma>2.5\gamma_{\rm crit}$):
\begin{enumerate}
\item \emph{I-J}: the exact one-step Jacobian, with $n_{1/2}=\ln2/(-\ln|\lambda_{\rm ret}|)$.
\item \emph{I-R1}: the rollout first crossing of the threshold $|\theta|\le\delta/2$.
\item \emph{I-R2}: the rollout last crossing of the threshold.
\item \emph{I-R3} (primary): the rollout envelope-rate fit over the late window, with $R^2\ge0.9955$ (median $0.9996$) in every cell.
\end{enumerate}

\begin{table}[!htb]\centering\small
\caption{The shape claims on all four instruments ($\delta=0.05$ rad, 48 $\gamma$ values, 3 emergent seeds). The rollout rate instrument reproduces the Jacobian; both threshold instruments fail below $\gamma_{\rm crit}$, so we never quote a rollout $n_{1/2}$ from a threshold there. Rows are stacked by quantity so that no column is truncated; no value differs from the per-instrument reading.}
\begin{tabular}{llcccc}
\toprule
Quantity & Instrument & Seed 42 & Seed 43 & Seed 44 & Mean $\pm$ sd\\
\midrule
$\gamma_{\rm crit}$ & --- & $0.02334$ & $0.01424$ & $0.02265$ & ---\\
\midrule
\multirow{4}{*}{Argmin$/\gamma_{\rm crit}$} & I-J & $0.8994$ & $0.9046$ & $0.9055$ & $0.9032\pm0.0027$\\
 & I-R1 & $0.0857$ & $0.1404$ & $0.0883$ & $0.1048\pm0.0252$\\
 & I-R2 & $0.2068$ & $0.1404$ & $0.3031$ & $0.2168\pm0.0668$\\
 & I-R3 & $0.8928$ & $0.9047$ & $0.9028$ & $0.9001\pm0.0052$\\
\midrule
\multirow{4}{*}{Slope Below} & I-J & $-1.0023$ & $-1.0016$ & $-1.0022$ & $-1.0020\pm0.0003$\\
 & I-R1 & $+0.0725$ & $+0.0688$ & $+0.0455$ & $+0.0623\pm0.0120$\\
 & I-R2 & $-1.0556$ & $+0.0688$ & $-0.8854$ & $-0.6240\pm0.4948$\\
 & I-R3 & $-0.9651$ & $-0.9977$ & $-0.9854$ & $-0.9827\pm0.0134$\\
\midrule
\multirow{4}{*}{Slope Above} & I-J & $+1.1262$ & $+1.1031$ & $+1.1254$ & $+1.1182\pm0.0107$\\
 & I-R1 & $+1.1027$ & $+1.0889$ & $+1.0977$ & $+1.0964\pm0.0057$\\
 & I-R2 & $+1.1027$ & $+1.0889$ & $+1.0977$ & $+1.0964\pm0.0057$\\
 & I-R3 & $+1.1255$ & $+1.1030$ & $+1.1231$ & $+1.1172\pm0.0101$\\
\bottomrule
\end{tabular}
\end{table}

The two instruments differ in level, not in rate. On the overdamped branch ($\gamma\ge2\gamma_{\rm crit}$), the decay \emph{rates} agree, while the \emph{threshold} instrument is offset by a per-checkpoint constant that is flat in $\gamma$. The full-window value for seed 44 is $0.9216\pm0.0027$, $0.0004$ below its pre-registered band, because a second, faster mode is still present in the window. The cause of the threshold offset is confirmed: the ring write deposits only a fraction of its amplitude in the slow branch (envelope-fit intercept over $\delta$: $0.638/0.644/0.327$).

\begin{table}[!htb]\centering\small
\caption{The instrument gap is a level, not a rate: on the overdamped branch ($\gamma\ge2\gamma_{\rm crit}$) the decay \emph{rates} agree while the \emph{threshold} instrument is offset by a per-checkpoint constant flat in $\gamma$. Seed 44's full-window value ($0.9216\pm0.0027$) sits $0.0004$ below its pre-registered band because that window still contains a second, faster mode (early-window ratio $0.539$); the window-split diagnostic converges monotonically to $1$ into the asymptote. Cause of the threshold offset, measured: the ring write deposits only a fraction of its amplitude in the slow branch (envelope-fit intercept over $\delta$ = $0.638/0.644/0.327$).}
\begin{tabular}{p{0.06\linewidth}p{0.25\linewidth}p{0.22\linewidth}p{0.30\linewidth}}
\toprule
seed & $\Gamma_{\rm jac}/\Gamma_{\rm R3}$ (asymptotic window) & $n_{\rm jac}/n_{\rm R1}$ (threshold) & CV of $n_{\rm jac}/n_{\rm R1}$ over the grid\\
\midrule
42 & $0.9935$ & $1.3307$ & \multirow{3}{*}{$2.2$--$3.7\%$ over 21--25 $\gamma$-points}\\
43 & $1.0000$ & $1.7981$ & \\
44 & $0.9948$ & $1.8204$ & \\
\midrule
mean & $0.995\pm0.003$ & per seed, at $\delta=0.05$ rad & $|d\log(\text{ratio})/d\log\gamma|\le0.0034$ (threshold $0.10$)\\
\bottomrule
\end{tabular}
\end{table}

We also vary the write amplitude. At $\delta=0.05/0.2/0.5$ rad, the I-R3 argmin is $0.8928/0.9007/0.8968$ (seed 42), $0.9047/0.9045/0.9174$ (seed 43), and $0.9028/0.9037/0.9070$ (seed 44). Over this order-of-magnitude ($10\times$) range the argmin shifts by at most $+1.4\%$, and the slopes change by less than $1.9\%$. At $\delta=0.5$ every run relaxed to baseline ($|\theta_{\rm final}|\le6.5\times10^{-10}$): no seed's write cleared its washboard barrier.

On the left branch, $\gamma<\gamma_{\rm crit}$, the coset eigenpair is complex with $|\lambda|^2=1-\gamma$. Then $n_{1/2}=\ln2/(-\tfrac12\ln(1-\gamma))$ is independent of both $\mu^2$ and the checkpoint. At $\gamma=0.002$, for instance, emergent seeds 42 ($\mu^2=5.449\times10^{-2}$) and 43 ($\mu^2=2.029\times10^{-2}$) return identical values: $\lambda_{\rm ret}=0.998999499$ and an $n_{1/2}$ of $692.5$, close to the predicted $2\ln2/0.002=693.1$.

\begin{figure}[htbp]
  \centering
  \includegraphics[width=\linewidth]{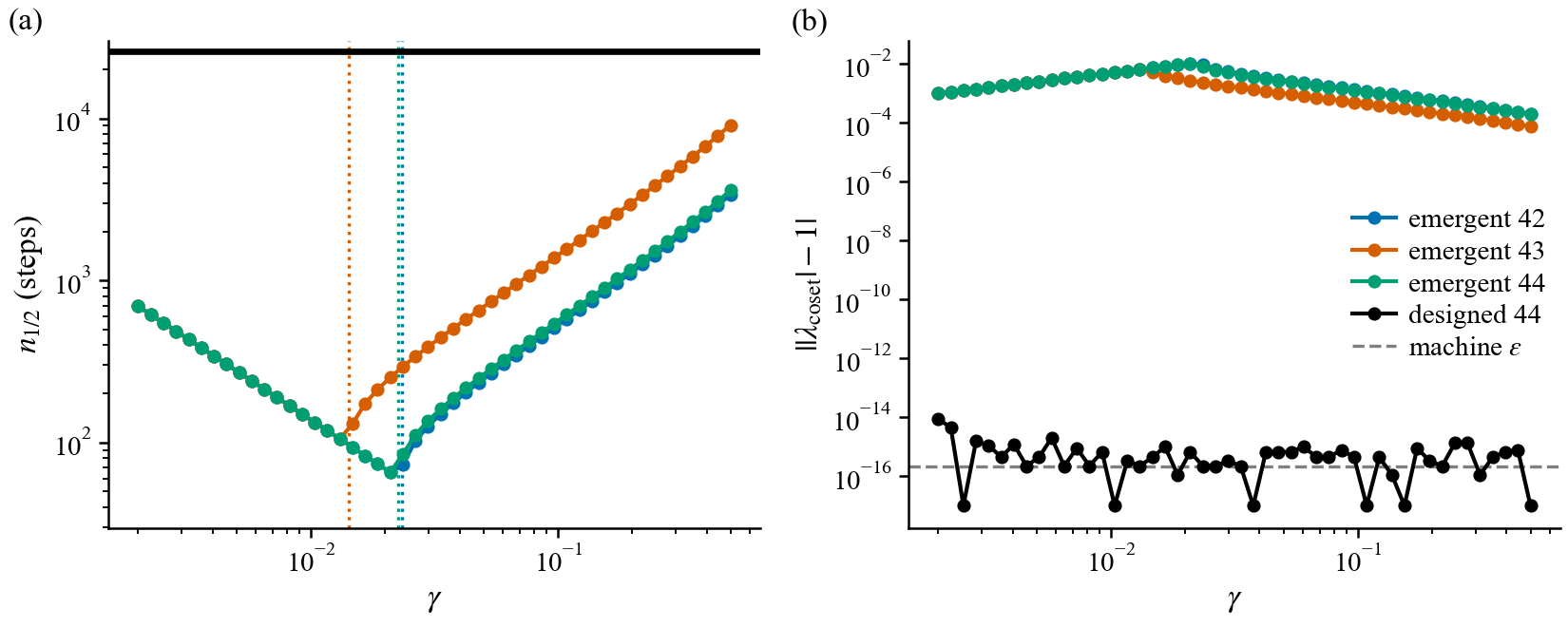}
  \caption{The emergent V-curve, un-collapsed, with the negative baseline beside it. (a) The $T=0$ half-life against friction for each emergent seed, from the one-step Jacobian; dotted lines mark each seed's $\gamma_{\rm crit}$. The black line at the top stands for the designed control's infinite half-life ($n_{1/2}=\infty$ at every $\gamma$); it is drawn at the axis top and is not a measured value. (b) $\big||\lambda_{\rm coset}|-1\big|$: it sits on the machine-$\varepsilon$ floor (dashed) for the designed symmetry but misses it by $\sim10^{-3}$ on all emergent seeds. The legend in (b) applies to both panels. \emph{Three emergent MLP seeds against one matched designed control.}}
  \label{fig:lambdacoset}
\end{figure}

\begin{figure}[htbp]\centering
\includegraphics[width=\linewidth]{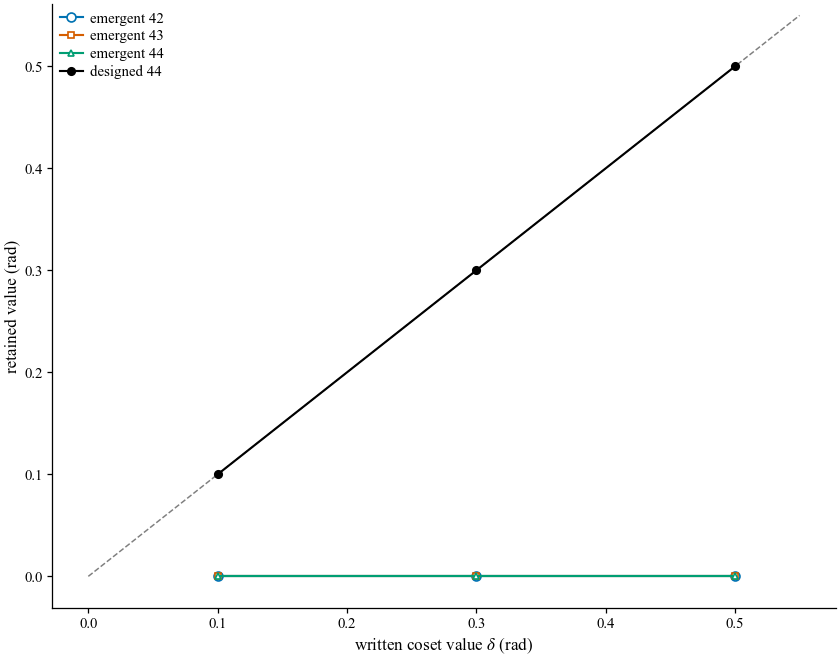}
\caption{The emergent unit has no continuous coset register. A written coset value $\delta$ is held for $20\,000$ steps at $T=0$, $\gamma=0.05$: on every emergent seed the value relaxes to the washboard minimum and nothing is retained, whereas the designed checkpoint returns $\delta$ exactly. The dashed diagonal is perfect retention. \emph{Three emergent MLP seeds plus a matched designed control: evidence.} This figure supports the capacity statement of $\approx1$--$1.6$ bits.}
\label{fig:registercapacity}
\end{figure}

\begin{figure}[htbp]\centering
\includegraphics[width=\linewidth]{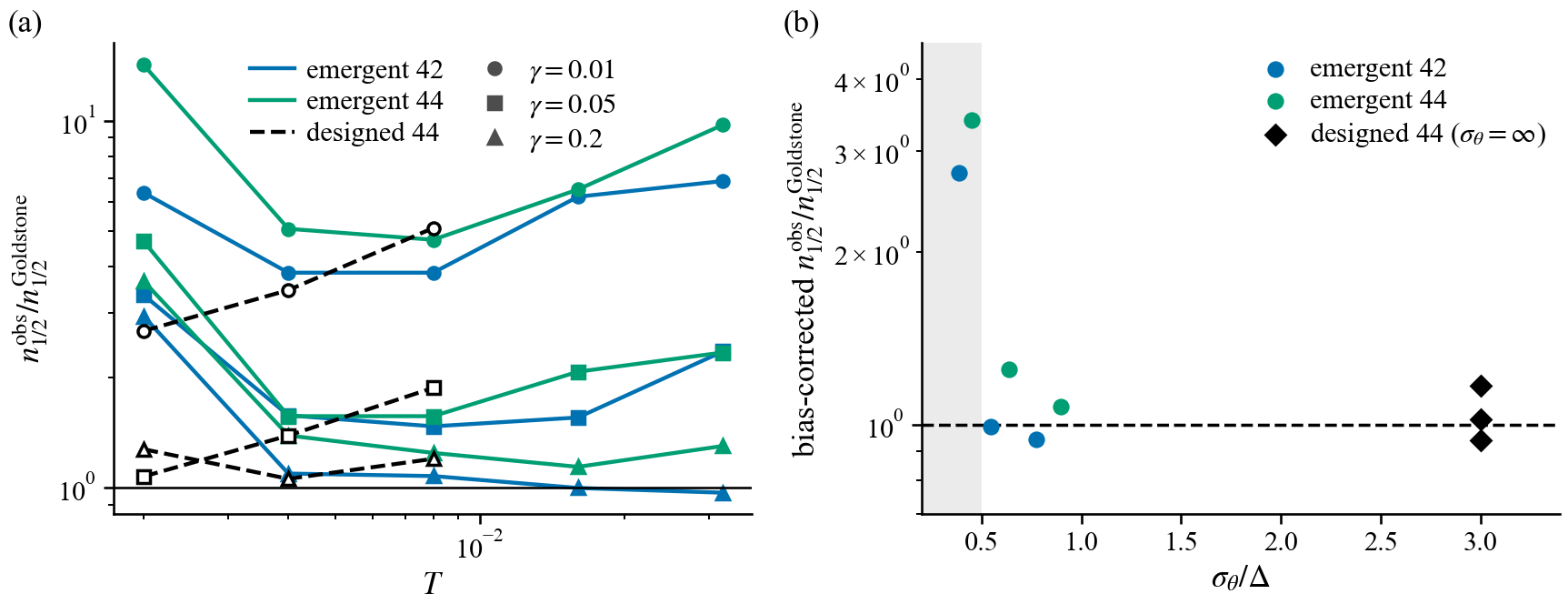}
\caption{The crossover $T^\star$, and why the raw curve cannot be read directly. (a) The raw excess over the exact-Goldstone law (solid line at $1$) is the same on the matched designed control (open markers, dashed) as on the emergent seeds (colors), so the excess is estimator bias rather than physics; marker shape gives $\gamma$. (b) After bias correction, the emergent coset joins the exact-Goldstone law (dashed) above $T^\star\approx3\times10^{-3}$; the shaded band marks the confinement-protected cells below it. The designed control has $\sigma_\theta=\infty$ and is drawn at the right edge. \emph{Multi-seed status: two emergent seeds and one matched designed control, an $n<3$ cell.} $\Delta=0.5$ rad, $\ell_\theta/\Delta<0.06$, \texttt{langevin\_noise="fdt"}, Newtonian kinetic mode: evidence.}
\label{fig:tstar}
\end{figure}

\section{The Friction-Hole Vault on the Emergent Arm}\label{app:vault}

Sec.~\ref{sec:vault} derives the absorb-only vault and reports the designed arm. Here we test the same predictions on the emergent arm.

On the emergent arm, the coset is bounded ($\mu^2_{\rm adiab}=2.77\times10^{-2}$ to $8.93\times10^{-2}$), so the block estimator of the designed arm is biased low by nearly $55\%$. We therefore replace it with a cross-validated, window-free Ornstein--Uhlenbeck fit, $\mathrm{MSD}(n)=A(1-e^{-n/\tau})$ with $\hat D=A/(2\varepsilon\tau)$. On the bounded cells, the law-referenced absolute vault is $106.1\pm5.0\times$ against the predicted $110.25\times$. The contrast metric, however, is native to the designed arm. Here the field-to-scalar control variance is $23.39\pm10.06$ against a pre-registered band of $[6.5,9.5]$, which triggers the falsifier: the control arm delocalizes while the field follows the absorb limit.

\begin{table}[!htb]\centering
\caption{The refrigerator law transfers to an emergent register: $\mathrm{Var}(p_i)/(M_iT)$ against the absorb-only prediction, 2048 walkers $\times$ 40 samples, 3 emergent seeds; over all 24 field cells observed/absorb $=0.9998\pm0.0019$, and at $\gamma_\phi=0.5$ $T_{\rm local}/T=0.12570$, a $7.955\times$ refrigerator against $7.942$ predicted. The coupled bath predicts $1.0$ everywhere and is measured at $0.2235$.}
\begin{tabular}{lcccccc}
\toprule
$T$ & $\gamma_\phi$ & $\gamma_{\rm eff}$ & absorb pred. & coupled pred. & measured & obs/absorb\\
\midrule
$4\times10^{-3}$ & $0.0$ & $0.050$ & $1.00000$ & $1.0$ & $0.99815\pm0.00257$ & $0.9981$\\
$4\times10^{-3}$ & $0.1$ & $0.145$ & $0.36249$ & $1.0$ & $0.36326\pm0.00067$ & $1.0021$\\
$4\times10^{-3}$ & $0.2$ & $0.240$ & $0.23082$ & $1.0$ & $0.23062\pm0.00080$ & $0.9991$\\
$4\times10^{-3}$ & $0.3$ & $0.335$ & $0.17480$ & $1.0$ & $0.17473\pm0.00046$ & $0.9996$\\
$4\times10^{-3}$ & $0.5$ & $0.525$ & $0.12591$ & $1.0$ & $0.12592\pm0.00055$ & $1.0001$\\
$8\times10^{-3}$ & $0.5$ & $0.525$ & $0.12591$ & $1.0$ & $0.12548\pm0.00032$ & $0.9966$\\
\bottomrule
\end{tabular}
\end{table}

\begin{table}[!htb]\centering\small
\caption{The hole confines an emergent register, and the laundering control shows that friction alone does not. $T=4\times10^{-3}$, 1024 walkers, burn-in $20\,000$--$80\,000$ steps, spread by interquartile range; ``hop'' means $|\theta|>1$ rad. The pre-registered stationary-spread comparison is \emph{void} here, and no in/out spread ratio may be quoted from these checkpoints: the control that must return exactly $1.000$ returns $0.4586\pm0.1181$.}\label{tab:vault-confine}
\begin{tabular}{llccccc}
\toprule
seed & arm & $\gamma_{\rm eff}$ & $\sigma_\theta$ (IQR) & hop fraction & Boltzmann at $T_{\rm local}$ & obs/pred\\
\midrule
42 & no hole, $\gamma=0.05$ & $0.050$ & $0.40795$ & $5.50\%$ & $0.24073$ & $1.6946$\\
42 & scalar $0.525$ (control) & $0.525$ & $0.20034$ & $0.73\%$ & $0.24073$ & $0.8322$\\
42 & $\gamma_\phi$ hole & $0.525$ & $0.06442$ & $0.0000$ & $0.08542$ & $0.7542$\\
43 & no hole & $0.050$ & $1.17601$ & $42.97\%$ & $0.46059$ & $2.5533$\\
43 & scalar $0.525$ & $0.525$ & $0.35335$ & $10.20\%$ & $0.46059$ & $0.7672$\\
43 & $\gamma_\phi$ hole & $0.525$ & $0.08628$ & $0.0000$ & $0.16343$ & $0.5279$\\
44 & no hole & $0.050$ & $0.36922$ & $2.36\%$ & $0.24034$ & $1.5363$\\
44 & scalar $0.525$ & $0.525$ & $0.21574$ & $0.26\%$ & $0.24034$ & $0.8977$\\
44 & $\gamma_\phi$ hole & $0.525$ & $0.07279$ & $0.0000$ & $0.08528$ & $0.8535$\\
\bottomrule
\end{tabular}
\end{table}

\begin{figure}[!htb]
\centering
\includegraphics[width=\linewidth]{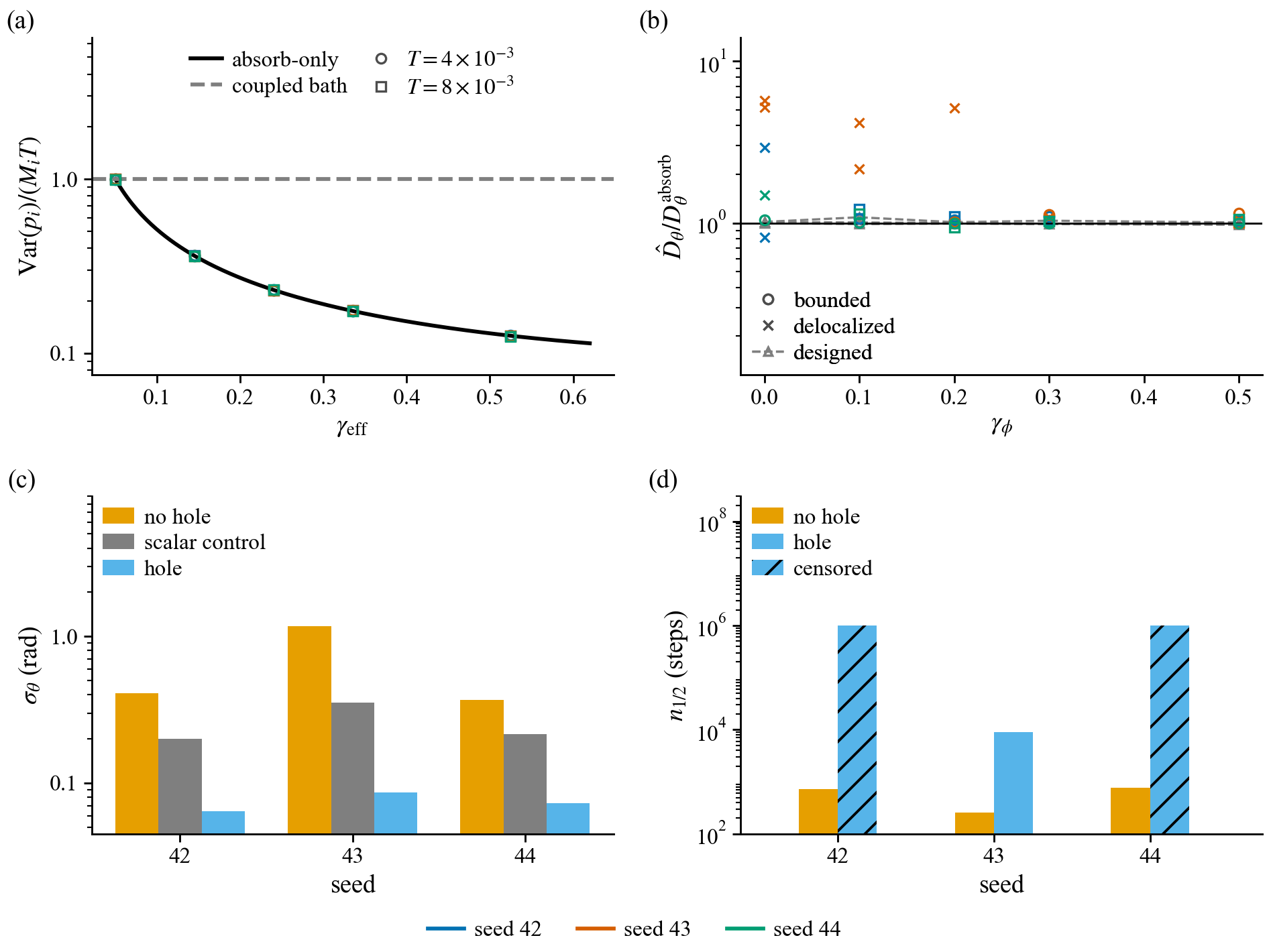}
\caption{The vault on an emergent register. Colors identify seeds (legend below the panels). (a) The refrigerator law, $\mathrm{Var}(p_i)/(M_iT)$ against $\gamma_{\rm eff}$ for 3 seeds $\times$ 2 temperatures (circles, squares), on the absorb-only curve (solid), with the coupled-bath prediction (dashed) rejected. (b) The $\gamma_{\rm eff}^{-2}$ diffusion law: $\hat D_\theta$ relative to the absorb-only law (line at $1$), with bounded cells as circles and squares, delocalized cells as crosses, and the designed cross-check (dashed) overlaid. (c) Confinement: stationary spread for no hole, the scalar control and the hole; the hop fractions are in Table~\ref{tab:vault-confine}. (d) Same-site first-passage half-life without and with the hole; hatched bars are censored lower bounds. \emph{(Emergent MLP, 3 seeds, $T\in\{4,8\}\times10^{-3}$, both above $T^\star$, $\Delta=0.5$ rad.)}}
\label{fig:vault-emergent}
\end{figure}

\end{document}